\documentclass[sigconf]{acmart}

\AtBeginDocument{%
  }

\usepackage{amsmath}
\usepackage{graphicx}
\usepackage{hyperref}
\usepackage{algorithm}
\usepackage{algpseudocode}
\usepackage{amsmath}
\usepackage{stmaryrd}
\usepackage{amsfonts}
\usepackage{array}
\usepackage{booktabs}
\usepackage{verbatim}
\usepackage{tikz}
\usetikzlibrary{arrows}
\usepackage{enumitem}

\usepackage{tikz}
\usepackage{booktabs}
\usepackage{multirow}
\usetikzlibrary{positioning, arrows.meta, shapes.geometric}

\setcopyright{none}

\begin{document}

\title{Multi-Behavior Sequential Modeling with Transition-Aware Graph Attention Network for E-Commerce Recommendation}

\author{Hanqi Jin}
\authornote{Equal contribution.}
\affiliation{%
  \institution{Taobao \& Tmall Group of Alibaba}
  \city{Beijing}
  \country{China}
}
\email{jinhanqi.jhq@alibaba-inc.com}

\author{Gaoming	Yang}
\authornotemark[1]
\author{Zhangming Chan}
\authornotemark[1]
\affiliation{%
  \institution{Taobao \& Tmall Group of Alibaba}
  \city{Beijing}
  \country{China}
}
\email{yanggaoming.ygm@taobao.com}
\email{zhangming.czm@taobao.com	}


\author{Yapeng Yuan}
\author{Longbin Li}
\affiliation{%
  \institution{Taobao \& Tmall Group of Alibaba}
  \city{Beijing}
  \country{China}
}
\email{yuanyapeng.yyp@taobao.com}
\email{lilongbin.llb@taobao.com}

\author{Fei Sun}
\affiliation{%
  \institution{University of Chinese Academy of Sciences	}
  \city{Beijing}
  \country{China}
}
\email{ofey.sunfei@gmail.com}

\author{Yeqiu Yang}
\author{Jian Wu}
\affiliation{%
  \institution{Taobao \& Tmall Group of Alibaba}
  \city{Beijing}
  \country{China}
}
\email{yangyeqiu.yyq@taobao.com}
\email{joshuawu.wujian@taobao.com}

\author{Yuning Jiang}
\author{Bo Zheng}
\affiliation{%
  \institution{Taobao \& Tmall Group of Alibaba}
  \city{Beijing}
  \country{China}
}
\email{mengzhu.jyn@taobao.com}
\email{bozheng@alibaba-inc.com}

\renewcommand{\shortauthors}{H. Jin et al.}
\newcommand{\hide}[1]{}

\begin{abstract}
User interactions on e-commerce platforms are inherently diverse, involving behaviors such as clicking, favoriting, adding to cart, and purchasing. The transitions between these behaviors offer valuable insights into user-item interactions, serving as a key signal for understanding evolving preferences. Consequently, there is growing interest in leveraging multi-behavior data to better capture user intent. Recent studies have explored sequential modeling of multi-behavior data, many relying on transformer-based architectures with polynomial time complexity. While effective, these approaches often incur high computational costs, limiting their applicability in large-scale industrial systems with long user sequences. 

To address this challenge, we propose the Transition-Aware Graph Attention Network (TGA), a linear-complexity approach for modeling multi-behavior transitions. Unlike traditional transformers that treat all behavior pairs equally, TGA constructs a structured sparse graph by identifying informative transitions from three perspectives: (a) \textbf{item-level transitions}, (b) \textbf{category-level transitions}, and (c) \textbf{neighbor-level transitions}. Built upon the structured graph, TGA employs a transition-aware graph Attention mechanism that jointly models user-item interactions and behavior transition types, enabling more accurate capture of sequential patterns while maintaining computational efficiency. Experiments show that TGA outperforms all state-of-the-art models while significantly reducing computational cost. Notably, TGA has been deployed in a large-scale industrial production environment, where it leads to impressive improvements in key business metrics. 
\end{abstract}


\maketitle

\section{Introduction}

Personalized recommender systems aim to infer users' preferences from their historical interactions and provide relevant item recommendations \cite{DBLP:journals/corr/HidasiKBT15,DBLP:conf/aaai/WuT0WXT19,DBLP:conf/cikm/SunLWPLOJ19,DBLP:conf/icde/XieSLWGZDC22,chang2023twin,si2024twin,zhou2018deep}. In e-commerce platforms, users engage with items through a variety of behavior types --- including clicks, add-to-cart actions, favorites, and purchases --- each reflecting different aspects of user intent and interest. These multi-behavior interactions offer rich signals for modeling evolving user preferences, where certain behaviors (e.g., purchases) indicate strong preference, while others (e.g., clicks) may reflect more exploratory or casual engagement.

Crucially, the transitions between these behaviors over time provide valuable insights into how users interact with items, offering a key signal for understanding dynamic user preferences. For instance, a user's progression from clicking an item to adding it to the cart and eventually purchasing it reveals a clear decision-making trajectory. This motivates recent efforts in incorporating such sequential behavioral transitions to gain deeper insights into user behavior and improve the accuracy of preference prediction.

A number of pioneering studies have explored various approaches for multi-behavior sequential modeling. Some works~\cite{DBLP:conf/mm/Su0LLLZ23,DBLP:conf/cikm/GuDWZLY20,DBLP:conf/www/WangZLLZLZ20,cho2023dynamic} divide multi-behavior sequences by behavior type and extract co-influence across behavior-specific sub-sequences. Others model item and behavior sequences independently, using behavior knowledge as auxiliary information to enhance representation learning~\cite{DBLP:conf/kdd/LiZLHMC18,DBLP:conf/www/TanjimSBHHM20,DBLP:conf/wsdm/ZhouDTY18,yan2024behavior}. Recently, several studies~\cite{DBLP:conf/kdd/YangHXLYL22,DBLP:conf/sigir/YuanG0GLT22} have introduced behavior-aware mechanisms that integrate contextual signals into item transitions to better capture evolving user interests.

However, most of these approaches rely on transformer-based architectures or other methods with polynomial time complexity, which often suffer from high computational costs — particularly when applied to long user behavior sequences. While these approaches achieve promising performance, their practical deployment in large-scale industrial systems remains challenging due to resource constraints. This creates a fundamental trade-off between model expressiveness and computational efficiency, especially in real-world scenarios where both accuracy and efficiency are critical.

\begin{figure}[t]
    \centering
    \includegraphics[width=0.8\columnwidth]{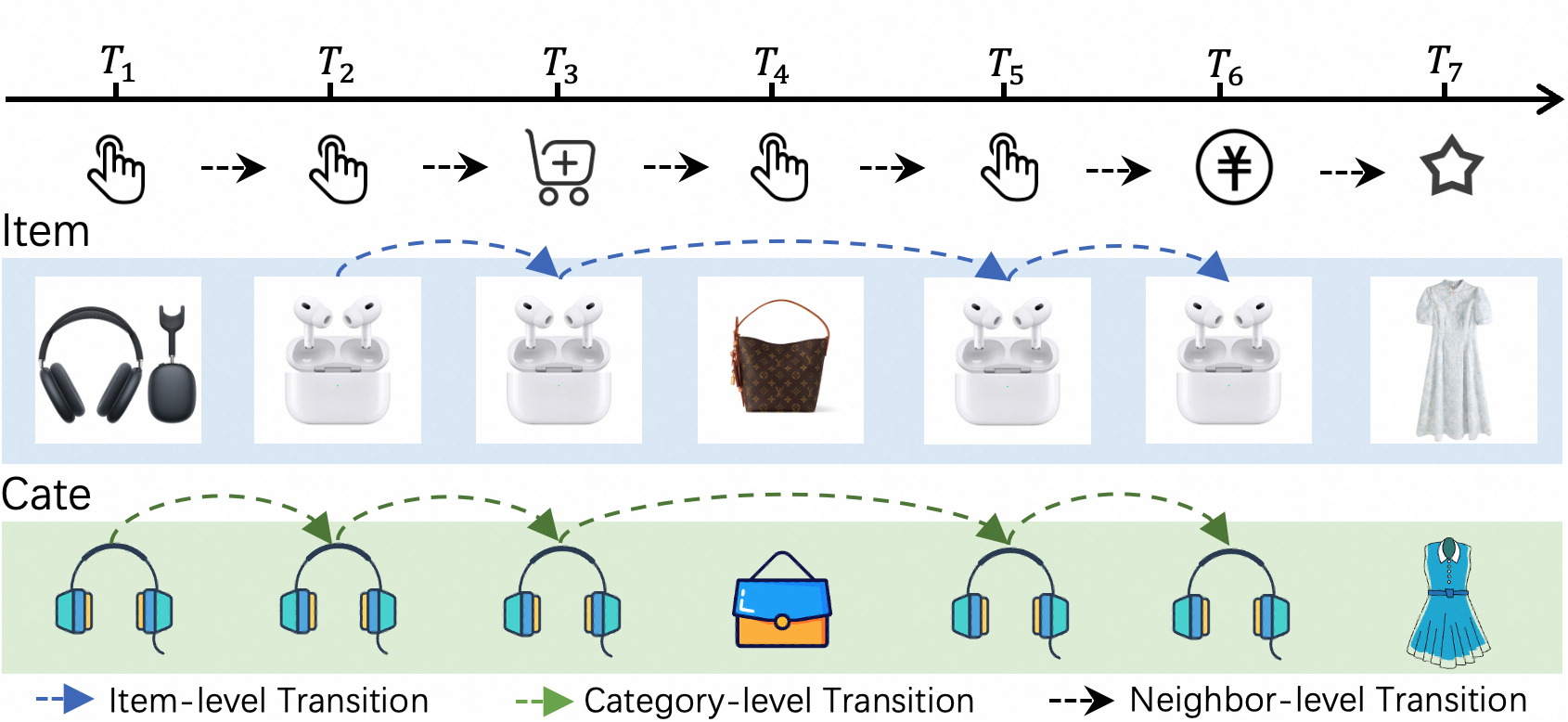} 
    \vspace{-2mm} 
    \caption{\small Item-level, category-level, and neighbor-level transitions in multi-behavior sequential modeling.} 
    \label{fig1} 
    \vspace{-6mm} 
\end{figure}

To address this challenge, we propose Transition-Aware Graph Attention Network (TGA), an efficient framework designed to explicitly model key behavioral transition patterns with linear time complexity. Unlike transformer-based approaches that rely on fully connected attention graphs with quadratic complexity, TGA constructs a structured sparse graph that identifies the most informative transitions across multiple relational views as shown in Figure \ref{fig1}: 
a.) \textbf{Item-level transitions}, capturing how users interact with the same item through multiple behavior types;
b.) \textbf{Category-level transitions}, modeling cross-item exploration within the same category;
c.) \textbf{Neighbor-level transitions}, representing local behavioral dependencies through temporal order.
Based on this structured graph, TGA introduces a novel Transition-Aware Graph Attention mechanism that jointly models user interactions and behavior transition types. By stacking multiple layers, TGA can capture high-order dependencies across multiple behavior dimensions. We summarize our main contributions as follows: 
\begin{itemize}[leftmargin=*] 
\item We propose TGA, an efficient framework that captures multi-behavior transitions with linear complexity, enabling its application to long user sequences in real-world scenarios.
\item We design a novel graph structure that captures informative behavior transitions from item-, category-, and neighbor-level perspectives, enabling fine-grained preference modeling. 
\item Experiments show that TGA achieves state-of-the-art performance with significantly lower resource cost, and it has been successfully deployed in a large-scale industrial environment. 

\end{itemize}

\section{METHODOLOGY}
\begin{figure}[t]
    \centering
    \includegraphics[width=0.85\columnwidth]{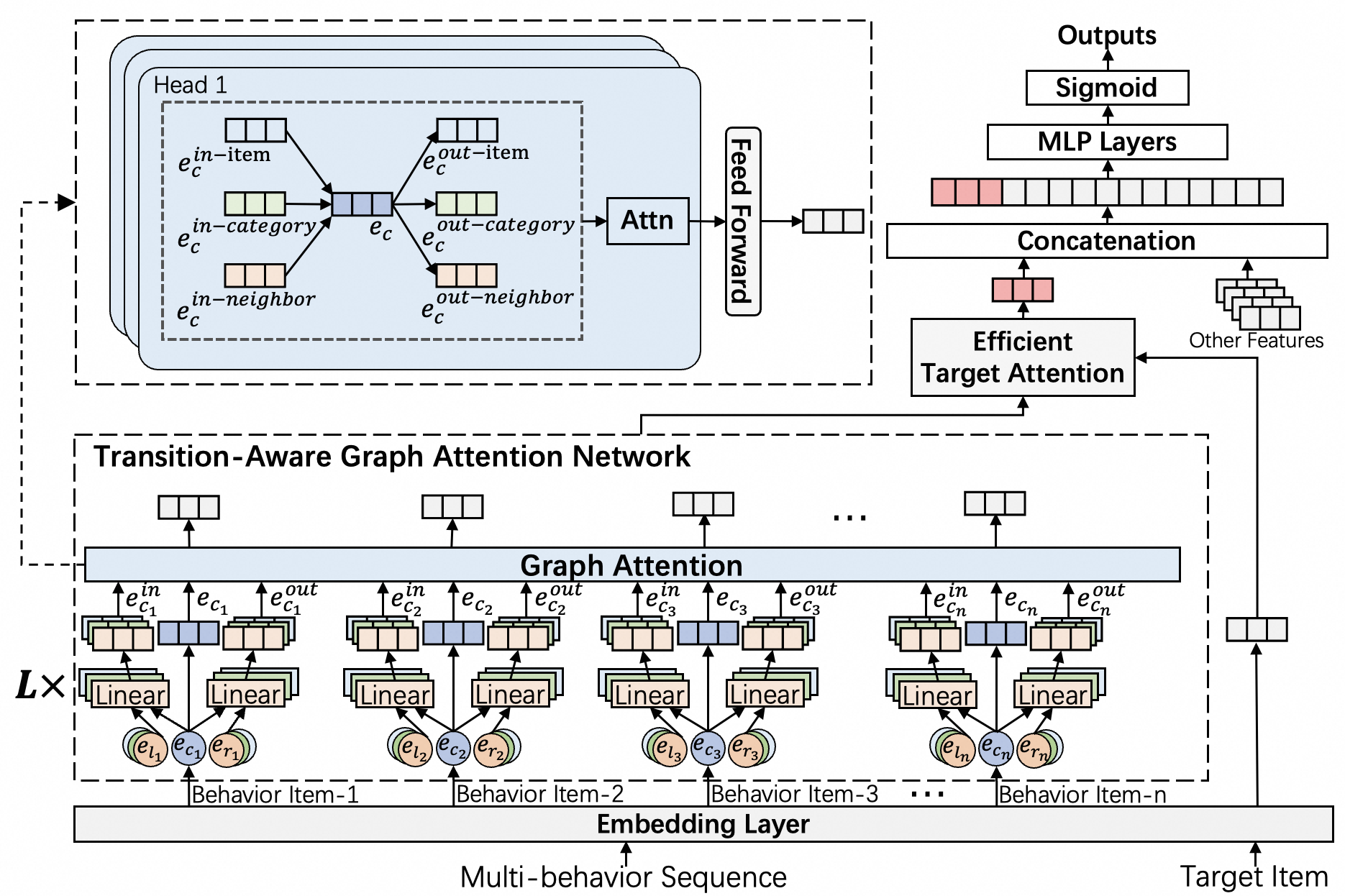}
    \vspace{-4mm} 
    \caption{\small The overview of our proposed TGA approach.}
    \vspace{-5mm} 
    \label{fig:learning_curves}
\end{figure}

As most multi-behavior sequential modeling approaches rely on transformer-based architectures with polynomial time complexity~\cite{DBLP:conf/mm/Su0LLLZ23,DBLP:conf/kdd/LiZLHMC18}, they often incur high computational costs, limiting their applicability in large-scale industrial systems with long user sequences. To address this limitation, we propose the \textbf{Transition-Aware Graph Attention Network (TGA)}, a computationally efficient framework that explicitly models key behavioral transition patterns through a structured sparse graph. 
As illustrated in Figure~\ref{fig:learning_curves}, we first apply the proposed TGA to the user behavior sequence $\mathcal{S}$, which effectively captures long-range dependencies and high-order interaction patterns.
The transformed sequence is then fed into the Efficient Target Attention to model interaction with the candidate item. The final prediction is obtained by feeding the user profile, the candidate item, and the attention output into a multi-layer perceptron with a sigmoid activation.

Before delving into the details of our proposed approach, we first formalize the problem setting. 
This work focuses on multi-behavior sequence modeling, evaluated on the conversion rate (CVR) prediction task. Let $\mathcal{D}(u_p, \mathcal{S}, c, y)$ denote the dataset, where $u_p$, $\mathcal{S}$, $c$, and $y$ represent the \textit{user profile}, \textit{user behavior sequence}, \textit{candidate item}, and \textit{conversion label}, respectively. Notably, each $\mathcal{S}$ contains \textit{long-term multi-type actions} (e.g., click, buy, add-to-cart, favorite). 
We design an effective and efficient multi-behavior sequence modeling method to $\mathcal{S}$ for better prediction performance. 

\subsection{Behavior Graph Construction}
\label{sec:graph_construction} 
In this section, we describe the graph construction process for the proposed TGA model.
We observe that user decision-making involves both fine-grained item-level interactions and broader category-level exploration patterns. For instance, high-cost items often trigger repeated interactions --- such as multiple revisits --- indicating a cautious evaluation process before purchase. At the category level, users frequently compare similar items within the same category (e.g., clothing), exhibiting behaviors such as repeated clicks and add-to-cart actions prior to final conversion.

Motivated by the above observation, we transform user multi-behavior sequences into a structured graph representation. This graph explicitly encodes user-item interactions, behavior types, and their transition relationships. Our construction process primarily revolves around three complementary perspectives:
\begin{itemize}[leftmargin=*]
\item \textbf{Item-Level Transition Edges.} These model user interactions on a specific item. If a user performs $b_x$ followed by $b_y$ on the same item $i$, we create a directed edge $(i, b_x) \rightarrow (i, b_y)$. 
\item \textbf{Category-Level Transition Edges.} These capture cross-item exploration within the same category. If a user transitions from behavior $b_x$ on item $i_x$ to $b_y$ on another item $i_y$ in the same category, we create a directed edge $(i_x, b_x) \rightarrow (i_y, b_y)$, modeling topical interest shifts (e.g., browsing different dresses). 
\item \textbf{Neighbor-Level Transition Edges.} These form the sequential backbone of the graph, connecting consecutive interactions in temporal order: $(i_x, b_x) \rightarrow (i_{x+1}, b_y)$, regardless of item or category, capturing raw sequences and local behavioral dynamics.
\end{itemize} 
\textbf{To maintain a sparse and efficient graph, we constrain each node to only its nearest temporal neighbors --- at most one predecessor and one successor within each relation view.} Higher-order dependencies are modeled by the network architecture. The average number of edges per node is 0.56 for item-level, 1.19 for category-level, and 2.00 for neighbor-level transitions.

\subsection{Transition-Aware Graph Attention Network}

Building upon the structured behavior graph, we propose a \textbf{Transition}\textbf{-Aware Graph Attention Network (TGA)} to model high-order dependencies among user interactions. Unlike conventional graph attention mechanisms that treat all edges uniformly, TGA explicitly distinguishes transitions across different behavioral relations (e.g., click$\rightarrow$purchase), thereby enabling fine-grained modeling of heterogeneous user behaviors.

\subsubsection{Node Representation} 
The model employs four distinct embedding tables for node representations in the graph. Specifically, each node $n$ (i.e., a behavior in user behavior sequence $\mathcal{S}$) is represented by concatenating the corresponding item embedding $e^{i}_n \in \mathbb{R}^{1 \times d}$, behavior embedding $e^{b}_n \in \mathbb{R}^{1 \times d}$, timestamp embedding $e^{t}_n \in \mathbb{R}^{1 \times d}$ and position embedding $e^p_n \in \mathbb{R}^{1 \times d}$ as $e_n = [e^{i}_n \oplus e^{b}_n \oplus e^{t}_n \oplus e^{p}_n]$, where $\oplus$ denotes concatenation operation. This setting enables our model to jointly capture item semantics, user behavior patterns, and temporal-spatial context information.

\subsubsection{Transition-Aware Graph Attention} 
To better capture contextual dependencies, we explicitly model directed transitions between nodes. Unlike most existing graph-based models~\cite{DBLP:conf/iclr/VelickovicCCRLB18}, which treat edges as undirected or ignore behavioral types, we design a learnable transformation mechanism for each type of transition edge. In our formulation, each node can act both as a sender and a receiver node in different transitions. For example, node $e_c$ participates in two directional transitions within each relationship view: an incoming transition ``$e_l \rightarrow e_c$'' and an outgoing transition ``$e_c \rightarrow e_r$''. 

Taking the incoming transition ``$ e_l \rightarrow e_c $'' as an example, let $b_l$ and $b_c$ denote the behavior type of nodes $e_l$ and $e_c$, respectively. 
We utilize both the information from the connected nodes and the transition edge to propagate meaningful features to the receiver node $e_c$. Specifically, we treat the node embeddings as input to a multi-layer perceptron (MLP), with the behavior transition information parameterizing the MLP weights and bias. More formally, for the given transition edge type $ b_l \rightarrow b_c $, we obtain the corresponding weight matrix $ \mathbf{W}^\text{in}_{b_l \rightarrow b_c} \in \mathbb{R}^{10d \times d} $ and bias vector $ \mathbf{b}^\text{in}_{b_l \rightarrow b_c} \in \mathbb{R}^{d} $. The output of this transformation is computed as: 
\begin{equation} 
\small 
\begin{aligned} 
e_c^\text{in} &= \textbf{W}^\text{in}_{b_l \rightarrow b_c} \cdot [e_l \oplus e_c \oplus (e^t_c - e^t_l) \oplus (e^p_c - e^p_l)] + \textbf{b}^\text{in}_{b_l \rightarrow b_c}, 
\end{aligned} 
\end{equation} 
where the temporal difference $(e^t_c - e^t_l)$ captures the time interval between two consecutive interactions, and the positional difference $(e^p_c - e^p_l)$ reflects their relative position in the behavior sequence. These temporal and relative position signals help the model better understand the dynamics and structure of user behavior. 
Similarly, for the outgoing transition $ e_c \rightarrow e_r $, we define the representation of sender node $e_c$ as 
\begin{equation} 
\small 
\begin{aligned} 
e_c^\text{out} &= \textbf{W}^\text{out}_{b_c \rightarrow b_r} \cdot [e_c \oplus e_r \oplus (e^t_r - e^t_c) \oplus (e^p_r - e^p_c)] + \textbf{b}^\text{out}_{b_c \rightarrow b_r}, 
\end{aligned} 
\end{equation}
where $ \mathbf{W}^\text{out}_{b_c \rightarrow b_r} \in \mathbb{R}^{10d \times d} $ and $ \mathbf{b}^\text{out}_{b_c \rightarrow b_r} \in \mathbb{R}^{d} $ are learnable parameters associated with the transition type $ b_c \rightarrow b_r $.

By applying such a bidirectional aggregation strategy across all three types of relationship views --- item-level, category-level, and neighbor-level transitions --- we obtain comprehensive transition-aware representations for each node. Specifically, the set of transition-aware representations for node $ e_c $ can be expressed as: 

\begin{equation} 
\small 
\begin{aligned} 
\mathcal{N}(e_c) = \bigcup_{v \in \{\text{item},\, \text{category},\, \text{neighbor}\}} \left\{ e^{\text{in-}v}_c,\, e^{\text{out-}v}_c \right\}. 
\end{aligned} 
\end{equation} 
This formulation enables the model to jointly capture both upstream and downstream behavioral dependencies within each transition view, thereby facilitating the learning of more comprehensive contextual patterns from the graph structure.

Then, we employ a multi-head graph attention mechanism to integrate all transition-aware representations in $\mathcal{N}(e_c)$. For each head, we compute the attention coefficients and aggregated representation as follows: 
\begin{equation} 
\small 
\begin{aligned} 
\hat{e}_c^{(k)} &= \sum_{u \in \mathcal{N}(e_c)} \alpha_k(u, e_c) \cdot \textbf{W}_k^{V} u, \\ 
\text{where \ }\alpha_k(u, e_c) &= \frac{\exp \left(\left(\textbf{W}_k^{K} u\right)^{\top} \cdot \textbf{W}_k^{Q} e_c\right)}{\sum_{z \in \mathcal{N}(e_c)} \exp \left(\left(\textbf{W}_k^{K} z\right)^{\top} \cdot \textbf{W}_k^{Q} e_c\right)}, 
\end{aligned} 
\end{equation} 
where each head $k$ learns independent transformations $W^{Q}_{k}, W^{K}_{k} \in \mathbb{R}^{d \times d_{k}}, W^{V}_{k} \in \mathbb{R}^{d \times d_{v}}$. 
The outputs from all heads are concatenated and linearly transformed to produce the final output $\hat{e}_c$. 

Finally, to further refine the output representations, we apply a feed-forward network (FFN) along with residual connections and layer normalization. The complete operation of the graph attention block is defined as follows:
\begin{equation}
\small 
\begin{aligned}
 e'_c =\text {LayerNorm}\left(e_c + \hat{e}_c\right), \tilde e_c=\text { LayerNorm }(e'_c + \mathrm{FFN}( e'_c)). 
\end{aligned} 
\end{equation} 
We refer to this entire process as a single-layer TGA module.

\subsubsection{Multi-Layer Extension for High-Order Dependency Modeling} 
As mentioned in Section~\ref{sec:graph_construction}, each node is only connected to its nearest temporal neighbors within each relation view. Therefore, the proposed TGA layer inherently models first-order relational dependencies within the behavior graph. To capture high-order interactions that span multiple steps in the user behavior sequence, we stack multiple TGA layers to form a deep architecture.

In this multi-layer design, the output representation of a node from the previous layer serves as the input to the next. This allows information to propagate across increasingly distant interactions through successive transformations. Formally, let $\mathcal{S}^l$ denote the representation of transformed sequence at the $l$-th layer, then: 
\begin{equation} 
\small 
\begin{aligned} 
\mathcal{S}^l &= \operatorname{TransGAT}^{(l)}(\mathcal{S}^{l-1}), 
\end{aligned} 
\end{equation} 
where $\operatorname{TranGAT}^{(l)}$ denotes the $l$-th layer. By stacking $L$ such layers, our model is able to encode progressively more abstract and long-range behavioral dependencies, effectively capturing high-order user interaction. The final representations are obtained from the topmost layer $L$, which consists of the transformed sequence $\mathcal{S}^L$, and are feeded into ETA module.


\subsubsection{Complexity Analysis}
Our complexity analysis focuses on two components: node-edge transformation and attention computation. 
Using weight matrices 
transformation incurs a time complexity of $O(N \cdot L \cdot d^2)$, where $L$ is the sequence length and $N$ is the number of layers.
For attention with six edges per node, all $N$ layers cost is $O(N \cdot L \cdot d)$.
Overall, our model achieves the linear time complexity of $O(N \cdot L \cdot d^2)$ with respect to sequence length
, which is significantly more efficient than the Transformer's $O(N \cdot L^2 \cdot d)$ when $L$ is large.

\begin{table}[t]
\caption{\small Performance comparison with baselines (``–'' = OOM);  ``*'' indicates TGA significantly better (p < 0.05).} 
\centering 
\resizebox{0.65\columnwidth}{!}{ 
\begin{tabular}{lcccc}
\hline
\multirow{2}{*}{Model} & \multicolumn{2}{c}{Taobao Dataset} & \multicolumn{2}{c}{Industrial Dataset} \\
\cline{2-5}
 & AUC & T\&I Speed & AUC & T\&I Speed \\
\hline
Transformer \cite{DBLP:conf/nips/VaswaniSPUJGKP17} & 0.7276* & 1.0x & - & -  \\
MB-STR \cite{DBLP:conf/sigir/YuanG0GLT22} & 0.7334*  & 0.8x & - & -  \\
END4Rec \cite{DBLP:conf/www/HanWWWLGLLC24} & 0.7405*  & 1.8x  & -  & -\\
Reformer \cite{DBLP:conf/iclr/KitaevKL20} & 0.7306*  & 1.7x & 0.8623* & 1.0x \\
Linear Transformer \cite{DBLP:conf/icml/KatharopoulosVMF20} & 0.7348*  & 1.9x & 0.8619* & 1.0x \\
Longformer \cite{DBLP:conf/emnlp/BeltagyPL20} & 0.7244* & 2.1x 
& 0.8617* & 1.2x\\
\midrule
TGA & \textbf{0.7454}  & \textbf{5.8x} & \textbf{0.8635}  & \textbf{3.4x}\\
\hline
\end{tabular}}
\label{table_result}
\end{table}




\begin{table}[t]
  \centering
  \caption{\small The results of ablation studies.} 
  \vspace{-3mm} 
  \label{table_ablations}
  \resizebox{0.7\columnwidth}{!}{ 
  \begin{tabular}{l|c|c} 
    \toprule
    Model & AUC & Improv.\\ 
    \midrule
    TGA & 0.8635  & - \\
    \quad w/o Item-Level Transitions     & 0.8618 & -0.0017  \\
    \quad w/o Category-Level Transitions & 0.8614 & -0.0021  \\
    \quad w/o Neighbor-Level Transitions & 0.8625 & -0.0010  \\
    \bottomrule
  \end{tabular}}
\end{table}

\section{EXPERIMENTS}
\subsection{Experimental Details}

\subsubsection{Dataset}
We evaluate our model on the post-click CVR prediction task 
on both public and real-world industrial datasets. For the public dataset, we use the Taobao dataset\cite{DBLP:conf/kdd/ZhuLZLHLG18}, which is widely used in previous work. For the industrial dataset, we use real-world data collected from the Taobao recommendation platform.

We split the Taobao dataset into training set (80\%), validation set (10\%), and test set (10\%) according to the timestep. 
The behavior sequence is truncated at length 256.
The industrial dataset comes from 14 days of user logs on a major e-commerce platform, with the next day's logs as the test set. The dataset contains over 10 billion instances, averaging 1,536 behavioral records per user. We use the most recent 1,024 interactions involving four behavior types (click, add-to-cart, favorite, purchase) as the multi-behavior sequence.

\subsubsection{Compared Methods.} 
To evaluate the effectiveness and efficiency of our approach, we compare it with the standard Transformer~\cite{DBLP:conf/nips/VaswaniSPUJGKP17} and efficient variants, i.e., Reformer~\cite{DBLP:conf/iclr/KitaevKL20}, Linear Transformer~\cite{DBLP:conf/icml/KatharopoulosVMF20}, and Longformer~\cite{DBLP:conf/emnlp/BeltagyPL20}. We also benchmark against state-of-the-art multi-behavior models, including END4Rec~\cite{DBLP:conf/www/HanWWWLGLLC24} and MB-STR~\cite{DBLP:conf/sigir/YuanG0GLT22}. For fair comparison, all models use the same input features and share a common architecture, differing only in the user multi-behavior modeling module. We set the embedding size $d$ to 64, batch size to 512, and number of layers $L$ to 3.

\subsubsection{Evaluation Metrics.} 
For offline evaluation, we adopt the widely used Area Under the Curve (AUC) and include Training \& Inference Speed (T\&I Speed) as a supplementary metric to measure model efficiency. 
\subsection{Experimental Results} 
\subsubsection{Overall performance}

As shown in Table~\ref{table_result}, TGA achieves an AUC of 0.7454 on Taobao Dataset, outperforming all baselines. More importantly, TGA is 5.8× faster in both training and inference than the standard Transformer, demonstrating the efficiency of our transition-aware graph attention mechanism. 

On Industrial Dataset, most existing methods either fail (e.g., Transformer). On those that can run on long sequences--Reformer, Linear Transformer, and Longformer---TGA still outperforms them in both accuracy and speed. 


\begin{figure}[t] 
    \centering 
    \includegraphics[width=0.7\columnwidth]{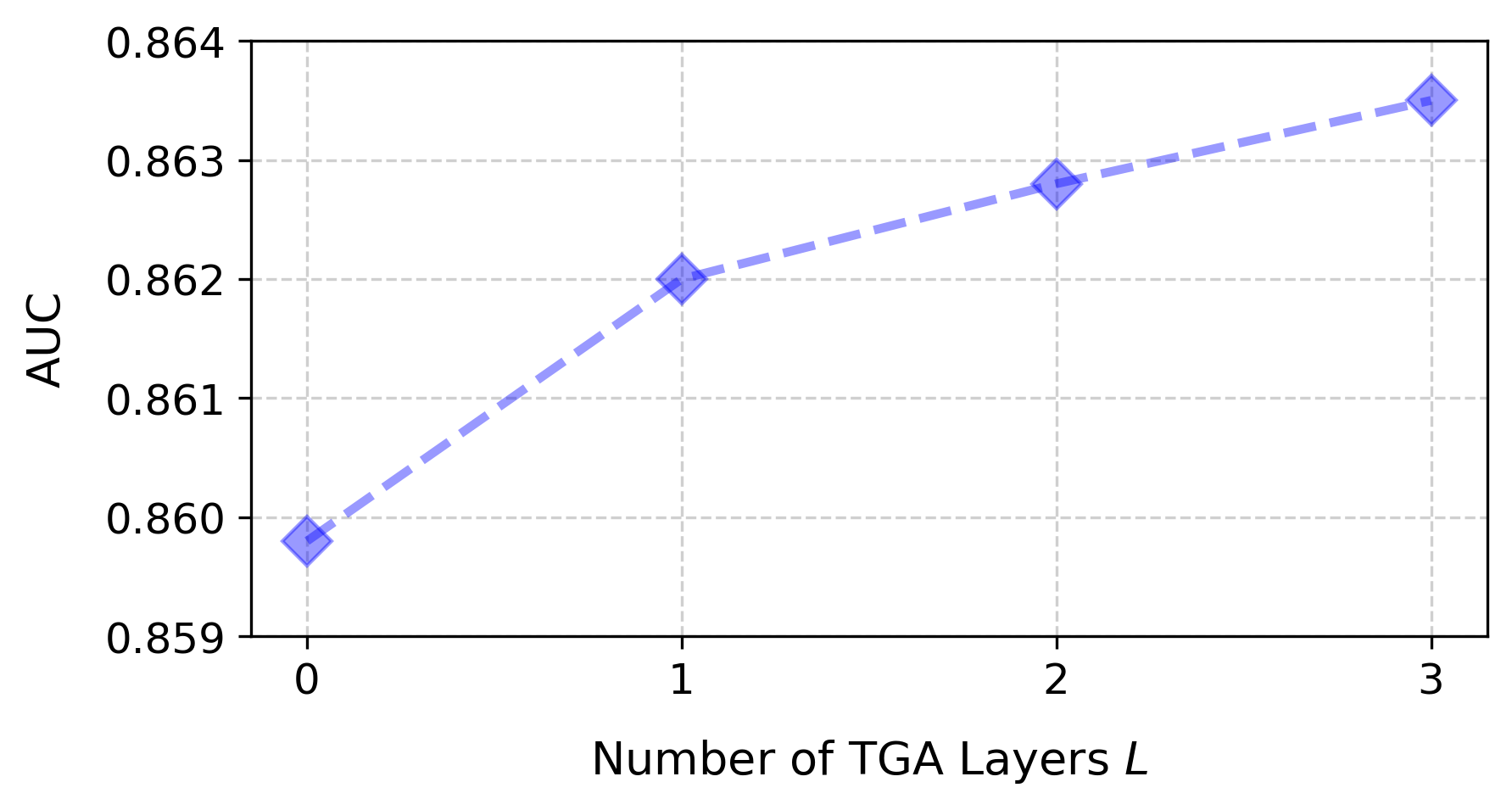} 
    \vspace{-3mm}
    \caption{\small AUC Performance with Increasing TGA Layers.} 
    \label{fig:tga_layers} 
    \vspace{-3mm} 
\end{figure} 

\subsubsection{Impact of TGA Transition Components} 
We examine the contribution of different components in the proposed TGA. Specifically, we remove each individual transition mechanism, i.e., item-level, category-level, and neighbor-level transitions within the structured graph, and evaluate their impact on model performance. As shown in Table~\ref{table_ablations}, removing any of these transitions leads to a performance decline, confirming the necessity of all components. Among them, the item-level and category-level transitions have the most significant impact, i.e., -0.0017 and -0.0021, indicating that fine-grained item interactions and category-based user behaviors provide the most informative signals for modeling conversion patterns.

\subsubsection{Impact of TGA Layer Depth} 
We investigate how the number of stacked TGA layers impacts model performance. As shown in Figure~\ref{fig:tga_layers}, with L layers, each node can aggregate information from its $L$-hop neighbors within the behavior-aware graph. The results demonstrate that increasing the number of layers leads to consistent performance improvements, validating the effectiveness of deep stacking in capturing higher-order user behavior dependencies.

\subsubsection{Online A/B Testing} 
We conduct online A/B testing in a real-world production environment to evaluate the effectiveness of TGA. The baseline model separately models individual behavior types (click, purchase, favorite, add-to-cart) without capturing cross-behavior patterns through multi-behavior sequence modeling. The experiment spans seven consecutive days and covers approximately 10\% of the platform's traffic. TGA achieves a 1.29\% improvement in CVR, along with an average increase of 1.79\% in GMV. These results demonstrate that TGA significantly outperforms the existing production model and delivers substantial business impact, especially considering the platform’s massive daily user volume.

\section{CONCLUSIONS}
In this paper, we proposed the TGA, an efficient multi-behavior sequential recommendation model tailored for long user interaction sequences in e-commerce. TGA models item-level, category-level, and neighbor-level transitions through a structured graph and a novel transition-aware graph attention mechanism, achieving linear time complexity with respect to sequence length. Experiments show that TGA achieves performance comparable to state-of-the-art methods while offering significantly higher efficiency. It has been deployed in a large-scale industrial system, where it demonstrates measurable improvements in key business metrics, validating its effectiveness in real-world applications.


\bibliographystyle{ACM-Reference-Format}
\bibliography{sample-base}


\end{document}